\documentclass{article}

\IfFileExists{arxiv.sty}{\usepackage{arxiv}}{\usepackage[margin=1in]{geometry}}

\usepackage[utf8]{inputenc}
\usepackage[T1]{fontenc}
\usepackage{hyperref}
\usepackage{url}
\usepackage{booktabs}
\usepackage{amsfonts}
\usepackage{amsmath}
\usepackage{amssymb}
\usepackage{mathtools}
\usepackage{microtype}
\usepackage[nameinlink,capitalise]{cleveref}
\usepackage{graphicx}
\usepackage[numbers,sort&compress]{natbib}
\usepackage{doi}
\usepackage{multirow}
\usepackage{array}
\usepackage{xcolor}
\usepackage{enumitem}

\graphicspath{{figures/}}

\providecommand{\shorttitle}{}
\renewcommand{\shorttitle}{Multi-Source Cognitive Taskonomy}

\newcommand{\safeincludegraphics}[2][]{%
  \IfFileExists{#2}{\includegraphics[#1]{#2}}{%
    \fbox{\parbox[c][4.6cm][c]{0.92\linewidth}{%
      \centering Missing figure: \texttt{\detokenize{#2}}}}%
  }%
}

\title{Beyond Single-Source Cognitive Taskonomy:\\
Multi-Source Task Relations through fMRI Transfer Learning}

\date{}

\author{Junfeng Xia\textsuperscript{*} \quad Wendu Li \quad Mengjiao Zhang \quad Jie Guo\\
\textit{Department of Biomedical Engineering, Southern University of Science and Technology}\\
Shenzhen, China\\
\textsuperscript{*}Corresponding author: \texttt{maoofiu@gmail.com}}

\hypersetup{
  pdftitle={Beyond Single-Source Cognitive Taskonomy: Multi-Source Task Relations through fMRI Transfer Learning},
  pdfsubject={fMRI, transfer learning, masked reconstruction, cognitive taskonomy},
  pdfauthor={Junfeng Xia, Wendu Li, Mengjiao Zhang, Jie Guo},
  pdfkeywords={fMRI taskonomy, masked reconstruction, multi-source transfer},
  colorlinks=true,
  linkcolor=blue,
  citecolor=blue,
  urlcolor=blue
}

\begin{document}
\maketitle

\begin{abstract}
Cognitive tasks are organized by both shared and specialized neural processes. Masked fMRI reconstruction provides a common self-supervised objective for quantifying transfer relations among cognitive task states, yet existing reconstruction-based taskonomies have focused primarily on one-to-one transfer from a single source task to a target. It therefore remains unclear how multiple source tasks jointly relate to the same target and which tasks should receive direct supervision when the available budget is limited. Here, we extend an fMRI cognitive taskonomy from single-source to multi-source transfer across 23 Human Connectome Project task states and use Boolean Integer Programming (BIP) to characterize budget-constrained task allocation. In total, we train 1,127 task-specific and transfer models. Single-source transfer is strongly directional and paradigm structured: motor states transfer well within the motor paradigm but provide limited support to most non-motor targets, consistent with a shared sensorimotor execution system overlaid by effector-specific representations. Multi-source transfer further depends on the specific composition of the source set, suggesting that many-to-one task relations are not fully captured by the pairwise taskonomy alone. Across supervision budgets, BIP repeatedly allocates direct supervision to several 0-back and 2-back working-memory states, although these states are not consistently the strongest individual sources. This pattern may reflect the joint engagement of frontoparietal control and category-selective visual systems in working-memory tasks. Together, these findings reveal a tightly coupled but cross-paradigm-limited motor cluster and working-memory states that occupy high-priority positions in global task allocation, consistent with their integration of perceptual, attentional, and executive processes. Our study extends reconstruction-based fMRI taskonomy from one-to-one transfer to many-to-one task relations and budget-constrained task dependencies.
\end{abstract}

\section{Introduction}
\label{sec:introduction}

Cognitive tasks are not isolated from one another.
A task state is typically composed of multiple shared cognitive processes and task-specific demands, such as perception, attention, memory, motor execution, and cognitive control, which are combined in different ways across tasks.
As a result, task relations may arise both within a paradigm, where tasks share similar demands, and across paradigms, where tasks rely on common cognitive components.
Cognitive ontologies have sought to formalize relations among mental constructs, experimental paradigms, and their neural implementations \citep{poldrack2011cognitiveatlas,turner2012cogpo}.
Large-scale modeling and neuroimaging studies further suggest that cognitive task representations exhibit structured organization rather than forming unrelated categories \citep{yang2019task,mensch2017learning,nakai2022representations}.
Characterizing this structure can help identify shared and task-specific neural computations and can also inform representation reuse when task-specific fMRI data are limited.

Transfer learning provides an operational way to quantify such task relations.
Rather than defining two tasks as related only because their activation maps or latent representations are similar, transfer analysis asks whether a representation learned from a source task can support low-data modeling of a target-task fMRI state.
This relation is inherently directional: the extent to which task $s$ supports task $t$ need not equal the extent to which task $t$ supports task $s$.
In computer vision, Taskonomy constructed directed task dependencies from transfer performance and further studied task-selection policies under limited supervision budgets \citep{zamir2018taskonomy}.
Subsequent work has examined transferability estimation, task cooperation and competition, and efficient task grouping from the perspectives of representation similarity and multi-task learning \citep{dwivedi2019representation,standley2020tasks,fifty2021groupings}.

In our prior work, we used masked fMRI reconstruction as a common transfer objective and constructed a single-source cognitive taskonomy across 23 Human Connectome Project (HCP) task states \citep{qu2024uncovering}.
That framework characterized one-to-one task relations of the form $s\rightarrow t$ and revealed coherent single-source transfer structure, including strong within-paradigm transfer among motor tasks and related structure among emotion, social cognition, and gambling tasks.
However, a one-to-one taskonomy only answers whether one source task can support one target task.
Because cognitive tasks are composed of multiple shared and task-specific components, a target task may benefit from several source-task representations simultaneously.
When multiple sources are used jointly, the basic unit of task relation changes from a single source--target edge to a source-set--target configuration, $S\rightarrow t$.
The transfer behavior of such a configuration may depend on how the cognitive components represented by different sources are combined, and therefore cannot be assumed to be fully described by a collection of pairwise transfer relations.

This many-to-one relation further raises a question about task supervision at the level of the entire task system.
Pairwise and source-set transfer distances indicate which representations can support a given target, but they do not directly determine which tasks should receive direct supervision when only a limited number of tasks can be selected.
A directly supervised task plays two roles: it is learned in its own right, and it becomes available as a source for supporting other targets.
Thus, the importance of a task under a supervision budget depends not only on local transfer strength but also on its coverage role in the global task system.
Budget-constrained Boolean Integer Programming (BIP) provides a way to convert measured transfer relations into a global task-allocation solution.

Based on this motivation, we extend reconstruction-based fMRI cognitive taskonomy from single-source transfer to multi-source transfer and use BIP to analyze task allocation under limited supervision.
Across 23 HCP task states, we train 1,127 task-specific and transfer models to characterize one-to-one task relations, many-to-one source-set relations, and budget-constrained supervision priorities.
The results show that motor tasks form a tightly connected within-paradigm cluster with relatively limited cross-paradigm support.
Multi-source transfer depends on the specific composition of the source set, indicating that many-to-one task relations are not fully captured by the pairwise taskonomy alone.
Several working-memory states, although not consistently the strongest individual sources, are repeatedly prioritized for direct supervision under different budgets.
Together, this study extends cognitive taskonomy from directed one-to-one transfer relations to many-to-one task relations and further reveals how local transfer structure gives rise to supervision priorities in the global task system.

\section{Related Work}
\label{sec:related_work}

\subsection{Cognitive Ontologies and Structured Task Representations}

The Cognitive Atlas and the Cognitive Paradigm Ontology provide formal frameworks for describing cognitive constructs, experimental paradigms, and their relationships
\citep{poldrack2011cognitiveatlas,turner2012cogpo}.
Beyond ontology construction, computational modeling and neuroimaging studies have shown that cognitive-task representations exhibit systematic organization.
Recurrent neural networks trained on multiple cognitive tasks develop representations that reflect task rules, functional specialization, and compositional structure
\citep{yang2019task}.
Likewise, representations learned across large collections of fMRI studies capture cognitive information shared across experimental conditions
\citep{mensch2017learning}.
An analysis of 103 cognitive tasks further showed that task-representation structure is broadly preserved across the cerebral cortex, cerebellum, and subcortical regions
\citep{nakai2022representations}.
Complementary optimization-based work has also examined how structural brain-network configurations vary across cognitive tasks
\citep{qu2024genetic}.
Together, these studies support the existence of a structured cognitive-task space.
However, they primarily characterize task relations through ontological links, representational geometry, or decoding performance, rather than directly measuring directed transfer between complete task states under a common low-data adaptation protocol.

\subsection{Transfer-Based Taskonomy and Multi-Task Relations}

Taskonomy quantifies task relations through source-to-target transfer performance and represents them as directed dependencies
\citep{zamir2018taskonomy}.
The original framework considers both single-source and source-set transfer, and connects these local transfer relations to supervision policies under a limited annotation budget.
It therefore provides a unified methodological framework for studying one-to-one transfer, many-to-one transfer, and budget-constrained task allocation.

Subsequent machine-learning studies have extended this direction from several perspectives.
Representation similarity has been used to estimate task transferability
\citep{dwivedi2019representation};
multi-task learning studies have examined which tasks should be learned jointly
\citep{standley2020tasks};
and efficient task-grouping methods have been proposed to reduce the cost of exhaustive combinatorial search
\citep{fifty2021groupings}.
These studies indicate that transfer performance depends not only on the source and target tasks individually, but also on the configuration of tasks used together.
Most existing evidence, however, is derived from computer-vision tasks and does not directly establish whether cognitive task states represented by distributed and temporally structured fMRI signals exhibit comparable transfer organization.

\subsection{Self-Supervised fMRI Representation Learning}

Masked autoencoding and masked image modeling established reconstruction of corrupted inputs as effective self-supervised objectives for representation learning
\citep{he2022mae,xie2022simmim}.
These objectives have recently motivated fMRI foundation models: large self-supervised models pretrained on broad collections of brain activity and reused as general representations for downstream decoding, prediction, or transfer.\citep{caro2024brainlm,xia2026brain,xia2026brainworld,qu2024uncovering,wang2025slim,wang2026omni,wang2026flexibrain}
BrainLM learns brain representations through masked signal prediction
\citep{caro2024brainlm};
Brain-JEPA predicts masked latent representations with spatiotemporal masking
\citep{dong2024brainjepa};
and Brain-DiT uses metadata-conditioned diffusion denoising to model heterogeneous brain states
\citet{xia2026brain}.
These approaches primarily aim to build general-purpose fMRI representations that can transfer across datasets, tasks, or downstream analyses.
In contrast, the present study uses task-specific self-supervised reconstruction as a controlled probe for comparing task-to-task transfer under a common objective and adaptation protocol, rather than proposing a new fMRI foundation model.

\subsection{Reconstruction-Based Cognitive Taskonomy}

The work most closely related to ours is our previous reconstruction-based cognitive taskonomy
\citet{qu2024uncovering}.
That study trained task-specific masked-reconstruction models and quantified directed transfer by freezing a source-task encoder and adapting a target decoder with limited target-task data.
This protocol characterized one-to-one relations across 23 HCP task states, but evaluated only one source task at a time.
It therefore did not capture many-to-one relations in which several source representations jointly support the same target, nor did it examine how the measured transfer structure could guide task allocation under a limited supervision budget.
The present study retains the same controlled reconstruction framework while extending the analysis to source-set transfer and budget-constrained allocation.

\section{Materials and Methods}
\label{sec:methods}

\subsection{Data and Task States}
\label{subsec:dataset}

We used task-fMRI data from the HCP S1200 release
\citep{vanessen2013hcp,barch2013function}.
The data were processed using the HCP minimal preprocessing pipeline
\citep{glasser2013preprocessing}
and parcellated into 360 cortical regions with the multimodal parcellation atlas
\citep{glasser2016mmp}.
Following \citet{qu2024uncovering}, each sample was represented as
\begin{equation}
    x\in\mathbb{R}^{R\times T},
    \qquad R=360,\quad T=20,
\end{equation}
where rows denote cortical regions and columns denote consecutive fMRI frames.
Participants were divided into training and test sets at a ratio of 4:1, with no subject overlap.

The analysis included 23 task states from seven HCP paradigms: eight working-memory states, five motor states, and two states each from emotion, gambling, language, social cognition, and relational processing
(\cref{tab:task_states}).

\begin{table}[t]
\caption{The 23 HCP task states used in the transfer analysis.}
\label{tab:task_states}
\centering
\small
\begin{tabular}{p{0.19\linewidth}p{0.66\linewidth}c}
\toprule
Paradigm & Task states & Number \\
\midrule
Working memory & 0-back body, faces, places, tools; 2-back body, faces, places, tools & 8 \\
Motor & left foot, right foot, left hand, right hand, tongue & 5 \\
Emotion & fear, neutral & 2 \\
Gambling & loss, win & 2 \\
Language & math, story & 2 \\
Social cognition & mental, random & 2 \\
Relational processing & match, relation & 2 \\
\bottomrule
\end{tabular}
\end{table}

\subsection{Masked fMRI Reconstruction}
\label{subsec:mae}

We adopted the masked-reconstruction architecture introduced in
\citet{qu2024uncovering}.
The ROI-by-time signal was divided into temporal patches and masked along the regional, temporal, or combined dimensions.
Patched signals were embedded by a one-dimensional convolution and processed by an eight-layer Transformer encoder and a six-layer Transformer decoder with sinusoidal positional encoding.
We used a temporal patch size of 2, a masking ratio of 50\%, and a hidden dimension of 1,024.
Models were optimized using AdamW with an initial learning rate of
$3\times10^{-5}$ and cosine annealing.

Let $\Omega$ denote the masked ROI--time entries.
The reconstruction loss was
\begin{equation}
\mathcal{L}_{\mathrm{rec}}
=
\frac{1}{|\Omega|}
\sum_{(r,\tau)\in\Omega}
\left(x_{r,\tau}-\hat{x}_{r,\tau}\right)^2.
\label{eq:reconstruction_loss}
\end{equation}
For each task state $t$, we trained a task-specific \emph{gold model} using its complete training partition and denoted its held-out loss by
$L_t^{\mathrm{gold}}$.
For the low-data validation analysis, a target-only \emph{fraction model} was trained using the same 1\% target-data budget used for transfer adaptation.

\subsection{Single- and Multi-Source Transfer}
\label{subsec:transfer}

Let $E_s$ denote the encoder of the gold model trained on source task $s$.
For any source set $S$ used to model target task $t$, we define the normalized transfer distance as
\begin{equation}
    d_t(S)
    =
    \frac{L_t(S)}{L_t^{\mathrm{gold}}},
\label{eq:transfer_distance}
\end{equation}
where $L_t(S)$ is the held-out reconstruction loss of the corresponding transfer model.
Lower values indicate stronger transfer relative to the intrinsic reconstruction difficulty of the target.

For single-source transfer, $S=\{s\}$.
The source encoder was frozen, and a target decoder was trained using 1\% of the target-task training data.
Evaluating all ordered pairs of distinct tasks produced
$23\times22=506$ single-source transfer models.

For multi-source transfer, the five strongest single-source candidates were retained for each target,
\begin{equation}
    \mathcal{C}_t=\{s_{t,1},\ldots,s_{t,5}\}.
\end{equation}
We evaluated every subset $S\subseteq\mathcal{C}_t$ with
$|S|\in\{2,3,4,5\}$.
Each frozen source encoder processed the same target sample, and the resulting representations were concatenated and projected before reconstruction:
\begin{equation}
    h_S(x)
    =
    P_{S,t}\!
    \left(
    \operatorname{Concat}_{s\in S}E_s(x)
    \right),
    \qquad
    \hat{x}=G_{S,t}\!\left(h_S(x)\right).
\label{eq:multisource_fusion}
\end{equation}
Only the projection $P_{S,t}$ and target decoder $G_{S,t}$ were optimized during adaptation.
This procedure yielded 598 multi-source models and 1,127 models in total, including the 23 gold models.

For visualization, transfer distances were standardized within each target and source-set cardinality:
\begin{equation}
    A_t^{(k)}(S)
    =
    -
    \frac{
    d_t(S)-\mu_t^{(k)}
    }{
    \sigma_t^{(k)}
    },
    \qquad |S|=k,
\label{eq:transfer_affinity}
\end{equation}
where $\mu_t^{(k)}$ and $\sigma_t^{(k)}$ were computed over all evaluated source sets of cardinality $k$ for target $t$.
The affinity was defined for $k\in\{1,2,3,4\}$.
For $k=5$, only one source set exists per target, so the corresponding models were reported using the unstandardized distance in \cref{eq:transfer_distance}.

Because candidate pools were target specific, multi-source affinities compare source compositions only within the same target and cardinality.
They do not define a common source axis across targets or constitute a formal measure of non-additive interaction.

\subsection{Budget-Constrained Task Allocation}
\label{subsec:bip}

We used Boolean Integer Programming (BIP) to determine which task states should receive direct supervision under a fixed budget
\citep{zamir2018taskonomy}.
Each evaluated transfer assignment was represented as a directed hyperedge
$e=(S_e\rightarrow t_e)$
with cost
\begin{equation}
    d_e=d_{t_e}(S_e).
\end{equation}
Let $z_s\in\{0,1\}$ indicate whether task $s$ receives direct supervision and
$y_e\in\{0,1\}$ indicate whether hyperedge $e$ is used to cover its target.
For supervision budget $B$, we solved
\begin{equation}
\begin{aligned}
\min_{z,y}\quad
& \sum_{e\in\mathcal{E}} d_e y_e,\\
\mathrm{s.t.}\quad
& z_t+\sum_{e:\,t_e=t}y_e=1,
&& \forall t\in\mathcal{V},\\
& y_e\le z_s,
&& \forall e\in\mathcal{E},\ \forall s\in S_e,\\
& \sum_{s\in\mathcal{V}}z_s\le B,\\
& z_s,y_e\in\{0,1\}.
\end{aligned}
\label{eq:bip}
\end{equation}
The first constraint requires each task to be either directly supervised or covered by one transfer assignment.
The second ensures that every source used by an assignment is directly supervised, and the third imposes the supervision budget.

We solved the BIP for
$B\in\{4,8,12,16\}$.
A task's \emph{allocation frequency} was defined as the number of evaluated BIP solutions in which it received direct supervision.
Because direct supervision both covers the task itself and makes it available as a source, this frequency is specific to the transfer costs, candidate pools, and budgets used in the optimization.

\section{Results}
\label{sec:results}

\subsection{Masked Reconstruction Provides a Common Transfer Substrate}
\label{subsec:results_representation}

\begin{figure}[t]
\centering
\safeincludegraphics[width=0.96\linewidth]{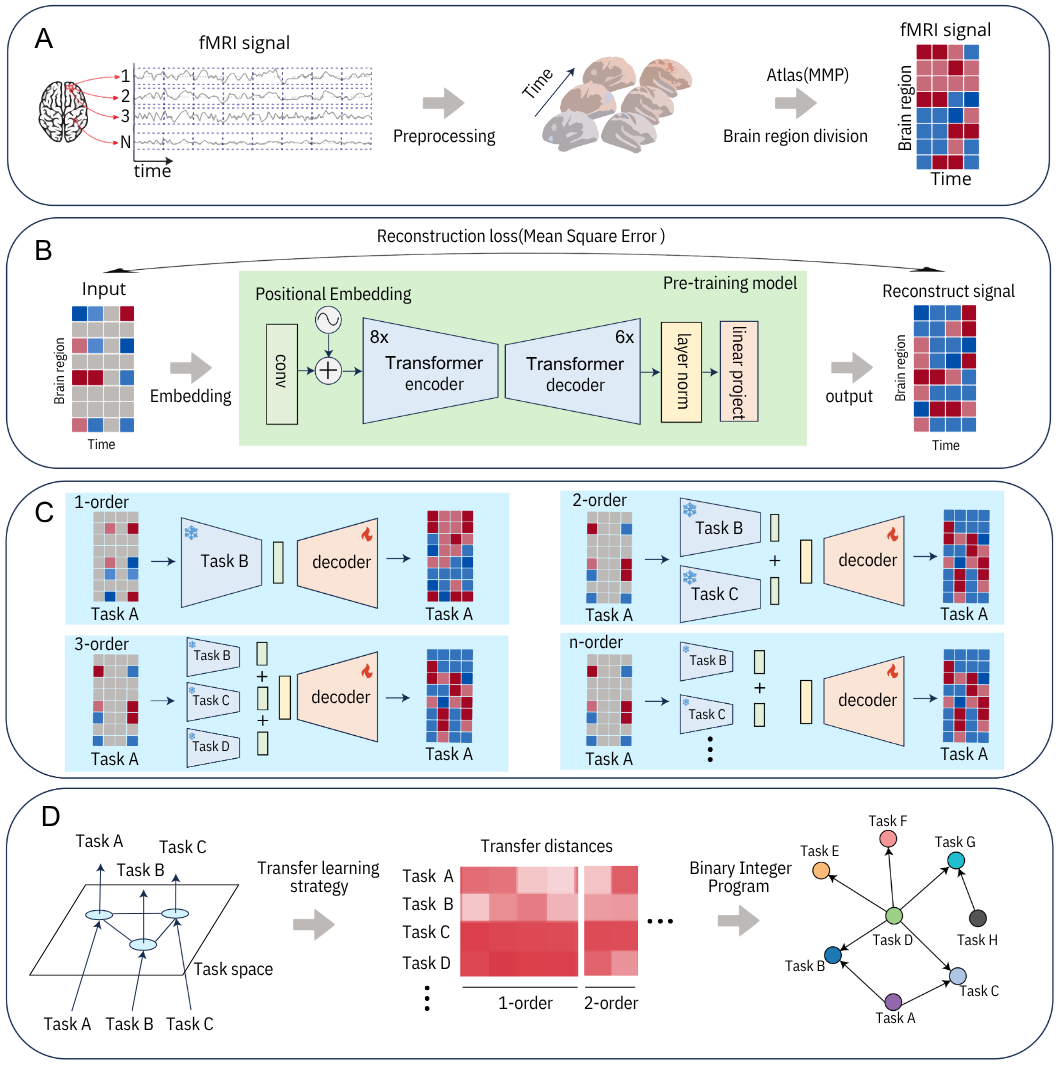}
\caption{Framework for multi-source cognitive taskonomy. (A) HCP fMRI preprocessing and MMP parcellation. (B) Masked reconstruction of ROI-by-time signals. (C) Single-source transfer uses one frozen source encoder; source-set cardinality $k$ uses $k$ frozen source encoders followed by a trainable projection and target decoder. (D) Normalized transfer distances define candidate assignments for budget-constrained BIP task allocation.}
\label{fig:framework}
\end{figure}

\Cref{fig:framework} summarizes the analysis.
HCP task-fMRI is converted to MMP ROI-by-time matrices, reconstructed with a masked Transformer, transferred through one or several frozen source encoders, and converted into a budget-constrained allocation problem.
Reconstruction supplies the same target objective for every source--target relation, allowing transfer models to be compared on a common held-out loss.

\begin{figure}[t]
\centering
\safeincludegraphics[width=0.96\linewidth]{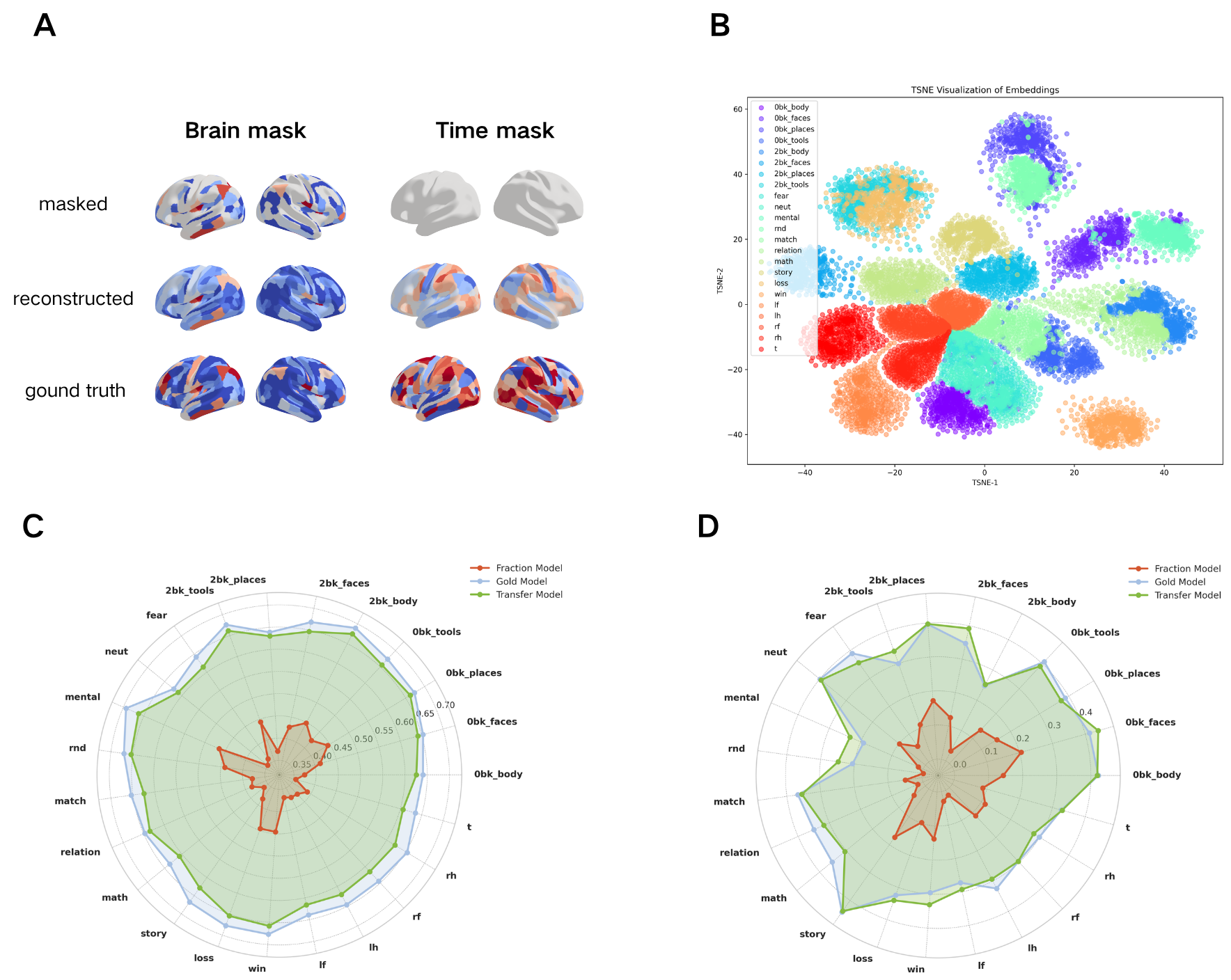}
\caption{Reconstruction and representation analyses. (A) Examples of brain-region and temporal masking, model reconstructions, and ground truth. (B) Two-dimensional t-SNE visualization of representations from 23 task states. (C) Reconstruction profiles for the low-data fraction model, full-data gold model, and transfer model. (D) Association between gradient-derived model connections and functional connectivity. Panels B and D are presented as descriptive diagnostics.}
\label{fig:representation}
\end{figure}

The validation analyses in \cref{fig:representation} reproduce the qualitative behavior established in the earlier reconstruction study \citet{qu2024uncovering}.
The model reconstructs both region-masked and time-masked signals, and the transfer model trained with minimal target supervision approaches the reconstruction profile of the full-data gold model more closely than the low-data target-only fraction model.
The t-SNE projection shows task-dependent organization of latent representations, while gradient-derived model connections covary with conventional functional connectivity.
These panels are retained as descriptive diagnostics of the reconstruction substrate and are not used to establish the multi-source or BIP conclusions.

\subsection{Single-Source Transfer Is Directed and Category Structured}
\label{subsec:results_first_order}

\begin{figure}[t]
\centering
\safeincludegraphics[width=0.98\linewidth]{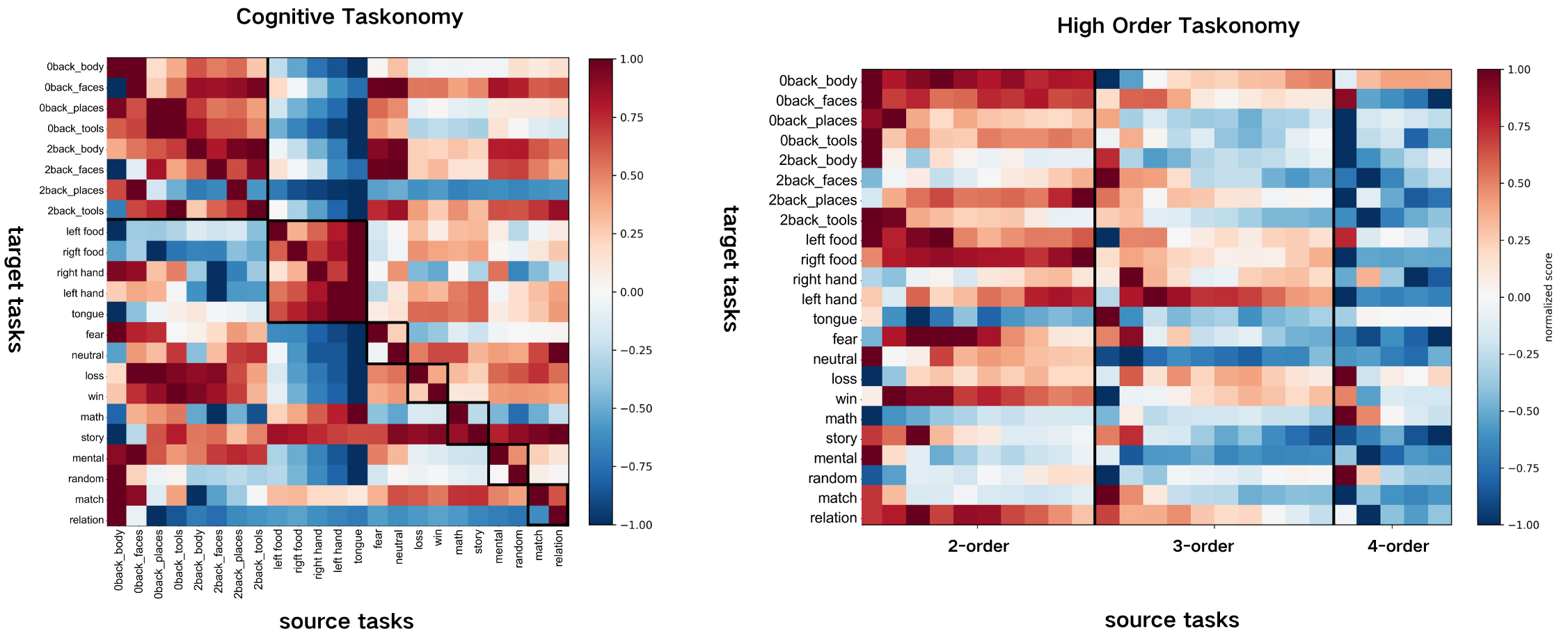}
\caption{Directed single-source taskonomy and multi-source transfer profiles. (A) Single-source affinity across 23 task states, with rows denoting targets and columns denoting sources; larger affinity corresponds to lower held-out transfer distance within a target. (B) Within-target standardized profiles for source-set cardinalities two, three, and four. Each row corresponds to a target, whereas columns index target-specific source combinations and therefore do not form a shared source axis across targets. Colors compare combinations only within the same target and cardinality. The all-five-source case is excluded because one source set per target does not permit within-target standardization.}
\label{fig:taskonomy}
\end{figure}

The single-source taskonomy in \cref{fig:taskonomy}A is not symmetric, confirming that incoming and outgoing transferability are distinct.
Consistent with the prior single-source analysis \citet{qu2024uncovering}, task states within the same paradigm often transfer more effectively than states from unrelated paradigms.
The five motor states form a coherent within-category block, particularly for homologous left--right effectors, but show weaker transfer to many non-motor targets.
This combination of strong internal transfer and weaker external transfer makes the motor family relatively specialized within the measured reconstruction framework.

Working-memory states require a directional interpretation.
As sources in the single-source analysis, they are not uniformly the strongest outgoing tasks; the earlier study found that several working-memory encoders transferred less broadly than sources such as math or neutral \citet{qu2024uncovering}.
As targets, working-memory states receive support from a wider range of non-motor sources in the current visualization.
Thus, target-side transferability should not be conflated with the claim that every working-memory state is a preferred individual source.

\subsection{Multi-Source Transfer Reveals Composition-Dependent Many-to-One Task Relations}
\label{subsec:results_multisource}

The multi-source analysis extends the transfer relation from a single source--target edge, ${s}\rightarrow t$, to a source-set--target configuration, $S\rightarrow t$.
Within each target-specific candidate pool, different source sets produced different held-out reconstruction distances even when the source-set cardinality was fixed.
Thus, the transfer relation of a target was determined not only by the number of source tasks, but also by which task representations were jointly used.

To relate these many-to-one relations to the pairwise taskonomy, we considered a simple pairwise prediction computed from the corresponding single-source transfer distances:
\begin{equation}
\hat d_t^{\mathrm{pair}}(S)
=
\frac{1}{|S|}
\sum_{s\in S} d_t({s}).
\label{eq:pairwise_prediction}
\end{equation}
This quantity represents the transfer distance expected if the performance of a source set were approximated by the average transfer strength of its individual members.
In contrast, the observed multi-source distance $d_t(S)$ is obtained by jointly applying all source encoders in $S$ and adapting the target reconstruction module.
The multi-source profiles in \cref{fig:taskonomy}B therefore directly measure many-to-one task relations that are not represented as explicit edges in the single-source taskonomy.

These profiles should be interpreted locally within the evaluated candidate pools.
Because candidate pools are target specific, columns in \cref{fig:taskonomy}B do not define a shared source axis across targets.
Moreover, different source-set cardinalities involve different numbers of source encoders and potentially different fusion capacity.
Accordingly, the result should be interpreted as composition-dependent many-to-one transfer within screened candidate pools, rather than as evidence of a formal non-additive interaction or a global higher-order taskonomy.

\subsection{Budget-Constrained Allocation Recurrently Selects Working-Memory States}
\label{subsec:results_bip}

The BIP analysis converts measured transfer distances into a budget-constrained task-allocation problem.
For each budget, a task can either receive direct supervision or be covered by a selected transfer assignment whose source tasks have been directly supervised.
Thus, the BIP solution reflects a global coverage objective over the task system rather than the strength of any single source--target edge.

To distinguish allocation priority from local pairwise transfer strength, we summarize each task's average single-source outgoing affinity as
\begin{equation}
\bar A_{\mathrm{out}}(s)
=
\frac{1}{|\mathcal{V}|-1}
\sum_{t\neq s} A_t^{(1)}({s}),
\label{eq:average_outgoing_affinity}
\end{equation}
where larger values indicate stronger average one-to-one transfer from task $s$ to other targets.
We also define the allocation frequency of task $s$ as
\begin{equation}
\mathrm{freq}(s)
=
\sum_{m\in\mathcal{M}} z_s^{(m)},
\label{eq:allocation_frequency}
\end{equation}
where $\mathcal{M}$ denotes the evaluated BIP solutions and $z_s^{(m)}=1$ indicates that task $s$ receives direct supervision in solution $m$.
The former quantity describes local outgoing transfer strength, whereas the latter describes how often a task is selected under the global budgeted allocation objective.

\begin{figure}[t]
\centering
\safeincludegraphics[width=0.98\linewidth]{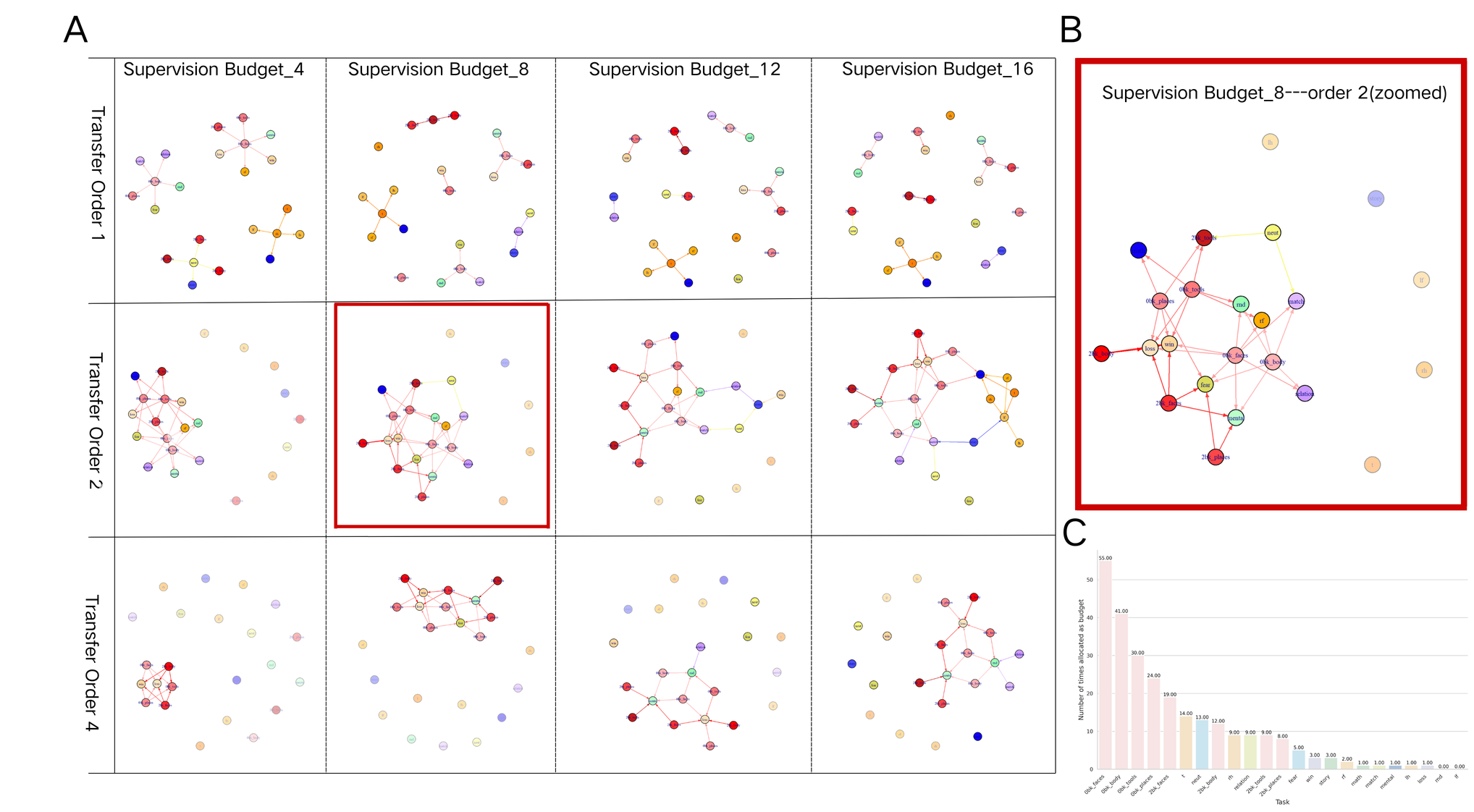}
\caption{Budget-constrained task allocation. (A) BIP solutions for supervision budgets 4, 8, 12, and 16 at the displayed source-set cardinalities. (B) Enlarged solution for budget 8 and two-source transfer. (C) Number of displayed solutions in which each task state is allocated direct supervision. These counts are descriptive outputs of the specified optimization problems and are not interpreted as intrinsic source quality or held-out policy performance.}
\label{fig:bip}
\end{figure}

The allocation frequencies in \cref{fig:bip}C are highest for several working-memory states, including 0-back faces, 0-back body, 0-back tools, 0-back places, and 2-back faces.
These states are repeatedly allocated direct supervision across the evaluated budgets, although working-memory states are not uniformly the strongest individual sources in the single-source taskonomy.
This pattern indicates that budget-constrained supervision priority is not equivalent to average pairwise source strength.
Instead, a task can be selected because it covers itself, enables low-cost transfer assignments to other targets, or fits the global coverage structure induced by the current cost matrix and budget.

The enlarged two-source, budget-eight solution in \cref{fig:bip}B illustrates this distinction.
Selected tasks form a non-uniform allocation graph in which directly supervised nodes both cover themselves and make particular transfer assignments feasible.
Therefore, the recurrent allocation of working-memory states should be interpreted as an outcome of the specified reconstruction distances, candidate pools, BIP constraints, and supervision budgets.
It does not imply intrinsic source quality, but it shows that local pairwise transfer and global task allocation capture different aspects of the learned task-relation structure.

\section{Discussion}
\label{sec}

This study extends reconstruction-based fMRI cognitive taskonomy from one-to-one task transfer to many-to-one source-set transfer and budget-constrained task allocation.
The main finding is that cognitive task relations are not fully described by a fixed pairwise transfer matrix.
Instead, the measured structure contains three linked levels: directed pairwise relations, composition-dependent source-set relations, and global supervision priorities induced by a budgeted allocation objective.
This organization reveals a coherent but cross-paradigm-limited motor cluster, source-set-dependent transfer to target states, and recurrent allocation of several working-memory states under supervision constraints.

\subsection{Motor Tasks Form a Coherent but Cross-Paradigm-Limited Cluster}

The single-source taskonomy shows that motor states transfer strongly within the motor paradigm but provide weaker support to many non-motor targets.
This pattern is consistent with the organization of the HCP motor task, in which different effectors share a common sensorimotor execution structure while retaining effector-specific representations.
Accordingly, motor states can be closely related to one another without becoming broadly useful sources for the full cognitive task set.
This distinction is important: strong within-paradigm transfer does not imply global transferability.
Rather, the motor family appears as a coherent task cluster whose representations are relatively specialized under the present reconstruction objective.

\subsection{Source-Set Transfer Captures Many-to-One Task Relations}

The multi-source analysis changes the unit of task relation from a single edge, ${s}\rightarrow t$, to a source-set--target configuration, $S\rightarrow t$.
Within a target-specific candidate pool, the measured transfer distance depends on which source representations are jointly used.
This shows that many-to-one task transfer should be measured directly, rather than treated as an explicit edge already present in the single-source taskonomy.

This conclusion does not require a claim of formal synergy or non-additive interaction.
Models with different source-set cardinalities receive different numbers of source encoders and may differ in fusion capacity.
Moreover, candidate pools are target specific, so the multi-source profiles do not define a shared global source axis across targets.
The appropriate interpretation is therefore narrower: source-set composition affects measured transfer within the evaluated candidate pools.
A formal interaction analysis would require defining a set utility and comparing observed source-set performance against an additive reference under matched capacity, data, and random-seed controls.

\subsection{Working-Memory States Receive High Priority in Budgeted Allocation}

The BIP results provide a complementary view of task relations.
Instead of asking whether one source supports one target, the allocation problem asks which tasks should receive direct supervision so that the remaining task states can be covered at low transfer cost.
A directly supervised task plays two roles: it is covered itself, and it becomes available as a source for other targets.
Therefore, allocation frequency is not equivalent to average pairwise outgoing transfer strength.

Several 0-back and 2-back working-memory states are repeatedly allocated direct supervision across budgets, even though working-memory states are not uniformly the strongest individual sources in the pairwise taskonomy.
This pattern suggests that these states occupy high-priority positions in the current global allocation problem.
One plausible interpretation is that HCP working-memory conditions combine multiple reusable components, including visual encoding of category-specific stimuli, attentional maintenance, comparison, decision making, and executive control
\citep{barch2013function,desposito2015working,duncan2010multiple,cai2024multidemand}.
The present analysis, however, does not isolate these mechanisms.
The recurrent allocation of working-memory states should therefore be interpreted as an optimization outcome under the specified reconstruction distances, candidate pools, BIP constraints, and budgets, rather than as evidence of intrinsic source quality.

\subsection{Implications for Budgeted fMRI Modeling}

The BIP formulation provides an explicit way to convert measured transfer distances into a task-allocation solution under a fixed supervision budget.
This is useful because task selection is otherwise often based on intuition, task labels, or pairwise similarity alone.
By making the cost matrix and constraints explicit, the formulation clarifies which task states are prioritized by the current reconstruction-based transfer structure.

At the same time, the current allocation results should be viewed as a model-based scenario analysis rather than a demonstrated deployment gain.
The selected task states have not been evaluated in newly trained downstream models or compared against random selection, greedy coverage, best-single-source ranking, same-paradigm selection, or pooled multi-task pretraining.
Thus, the practical contribution is the allocation framework and its observed task-priority pattern, not proof that the selected states universally improve data-efficient fMRI modeling.

\subsection{Limitations}

Several limitations define the scope of the conclusions.
First, this study extends a previously published single-source reconstruction framework, so the independent contribution lies in the multi-source transfer and budgeted allocation analyses.
Second, screening each target to its five strongest single-source candidates may exclude source sets whose members are weak individually but useful jointly.
Third, source-set cardinalities differ in the number of source encoders and potentially in fusion capacity, preventing a causal interpretation of cardinality itself.
Fourth, the current analysis does not estimate uncertainty across random seeds, alternative subject splits, or external datasets.
Fifth, BIP allocation frequencies are affected by category size, self-coverage, candidate screening, cost normalization, and the chosen budgets; they are not intrinsic source-quality scores.
Sixth, reconstruction MSE captures only one aspect of representation quality.
Finally, all analyses are based on one dataset and one cortical parcellation, and the participant-level split does not explicitly control family structure in HCP.

\section{Conclusion}
\label{sec}

This study extends reconstruction-based fMRI cognitive taskonomy from directed one-to-one transfer to many-to-one source-set transfer and budget-constrained task allocation.
Across 23 HCP task states and 1,127 models, single-source transfer reveals a directional and paradigm-structured taskonomy, with motor states forming a tightly connected but cross-paradigm-limited cluster.
Multi-source transfer further shows that target-state reconstruction depends on the composition of the source set, indicating that many-to-one task relations are not explicitly represented by pairwise transfer edges alone.
Finally, BIP allocation repeatedly assigns direct supervision to several working-memory states across budgets, separating local pairwise transfer strength from global supervision priority under the current objective.
Together, these results suggest that cognitive task relations in fMRI contain both specialized task clusters and budget-dependent task dependencies, providing a cautious framework for studying how learned task-state representations can be reused under limited supervision.



\begingroup
\raggedright
\sloppy
\bibliographystyle{unsrtnat}
\bibliography{references}
\endgroup

\end{document}